\documentclass{article}

\usepackage{PRIMEarxiv}

\usepackage[utf8]{inputenc} 
\usepackage[T1]{fontenc}    
\usepackage{hyperref}       
\usepackage{url}            
\usepackage{booktabs}       
\usepackage{amsfonts}       
\usepackage{nicefrac}       
\usepackage{microtype}      
\usepackage{lipsum}
\usepackage{fancyhdr}       
\usepackage{graphicx}       
\graphicspath{{media/}}     
\usepackage{multicol}
\usepackage{multirow}
\usepackage{amsmath}
\usepackage{subcaption,lipsum}
\title{CwA-T: A Channelwise AutoEncoder with Transformer for EEG Abnormality Detection
\thanks{\textit{\underline{Citation}}: 
\textbf{Under Review}} 
}

\author{
  Youshen Zhao \\
  Department of Informatics \\
  Kyushu University \\
  Fukuoka\\
  \texttt{ysZhaocs@gmail} \\
   \And
  Keiji Iramina \\
  Department of Informatics \\
  Kyushu University \\
  Fukuoka\\
}

\begin{document}
\maketitle

\begin{abstract}
Electroencephalogram (EEG) signals are critical for detecting abnormal brain activity, but their high dimensionality and complexity pose significant challenges for effective analysis. In this paper, we propose CwA-T, a novel framework that combines a channelwise CNN-based autoencoder with a single-head transformer classifier for efficient EEG abnormality detection. The channelwise autoencoder compresses raw EEG signals while preserving channel independence, reducing computational costs and retaining biologically meaningful features. The compressed representations are then fed into the transformer-based classifier, which efficiently models long-term dependencies to distinguish between normal and abnormal signals. Evaluated on the TUH Abnormal EEG Corpus, the proposed model achieves 85.0\% accuracy, 76.2\% sensitivity, and 91.2\% specificity at the per-case level, outperforming baseline models such as EEGNet, Deep4Conv, and FusionCNN. Furthermore, CwA-T requires only 202M FLOPs and 2.9M parameters, making it significantly more efficient than transformer-based alternatives. The framework retains interpretability through its channelwise design, demonstrating great potential for future applications in neuroscience research and clinical practice. The source code is available at \url{https://github.com/YossiZhao/CAE-T}.
\end{abstract}





\section{Introduction}
\label{introduction}
Brain disorders such as Alzheimer’s disease, epilepsy, Parkinson’s disease have attracted significant research interest due to their profound impact on patients’ quality of life and healthcare systems globally \cite{feigin2019global,feigin2021burden}. Timely and accurate diagnosis is crucial for effective intervention and management, necessitating reliable tools capable of capturing the dynamic changes in brain activity. Electroencephalography (EEG), a cost-effective and non-invasive method for real-time monitoring of brain function, has become a cornerstone in clinical practice for detecting brain disorders. By measuring electrical activity in the brain, EEG provides valuable insights into neural dynamics, particularly for conditions like epilepsy and Alzheimer’s disease, where the identification of abnormal patterns is critical for diagnosis and treatment. 

Recent advances in deep learning (DL) have significantly enhanced the capabilities of computer-aided diagnosis (CAD) systems for EEG analysis. These systems excel at extracting complex, high-dimensional features from raw EEG signals, improving diagnostic accuracy across various applications \cite{rivera2022diagnosis,lima2022comprehensive,dara2018feature}. However, the inherent challenges of analyzing long-term EEG signals—such as their high dimensionality and susceptibility to noise—necessitate novel approaches for robust and interpretable solutions.

DL-based CAD systems for EEG analysis can generally be categorized into two paradigms. The first involves preprocessing and feature extraction steps, such as filtering, artifact removal, and transformation into alternative domains (e.g., frequency or time-frequency domains using Fourier or wavelet transforms). These methods reduce noise and enhance clinically relevant features, particularly for noisy signals influenced by muscle artifacts or electrode interference \cite{amin2015feature,sun2023survey,hamad2016feature, panda2010classification}. For instance, Velasco et al. \cite{velasco2023motor} achieved 100\% classification accuracy in motor imagery EEG using a multivariate time-series approach combined with discrete wavelet transform (DWT). Similarly, Wang et al. \cite{wang2018epileptic} demonstrated a 98.1\% detection accuracy for epileptic seizures in long-term EEG recordings using wavelet decomposition and directed transfer functions.

The second paradigm bypasses preprocessing by directly inputting raw EEG data into DL architectures like convolutional neural networks (CNNs) or recurrent neural networks (RNNs). These models leverage their ability to learn intricate patterns directly from raw data, achieving remarkable results in various tasks \cite{song2022eeg,xie2022transformer,wei2023tc,lih2023epilepsynet}. For example, a CNN-LSTM model proposed by Xu et al. \cite{xu2020one} achieved 99.39\% accuracy in binary epileptic seizure classification. Similarly, RNN-based approaches have proven effective in tasks like schizophrenia detection, achieving up to 98\% accuracy \cite{supakar2022deep}.

In recent years, transformer-based architectures have emerged as a transformative innovation in DL, excelling in fields like natural language processing and computer vision. Their application to EEG analysis has yielded competitive or superior performance compared to traditional architectures \cite{song2022eeg,xie2022transformer,wei2023tc,lih2023epilepsynet}. Notable examples include transformer models for emotion recognition and epilepsy detection that achieve high accuracy by leveraging positional encodings and person-specific embeddings \cite{li2022eeg, lih2023epilepsynet}. However, transformers’ computational complexity and the “black-box” nature of their decision-making process pose significant challenges, particularly for applications requiring lightweight models and interpretability \cite{adadi2018peeking}.

This study introduces a novel framework that integrates a channelwise autoencoder with a lightweight, single-head transformer-based classifier for EEG abnormality detection. The framework seeks to address key challenges in EEG analysis by focusing on efficiency, interpretability, and competitive performance. Specifically:
\begin{enumerate}
    \item \textbf{Efficiency and practicality} The framework eliminates the need for complex preprocessing steps, enabling direct processing of long-term raw EEG signals while maintaining low computational costs.
    \item \textbf{Interpretability} The model preserves channel independence, aligning with the spatial structure of EEG signals. This design enhances biological interpretability and lays a foundation for future neuroscience research and clinical applications.
    \item \textbf{Performance} The proposed method achieves competitive classification accuracy while significantly reducing computational costs compared to standalone transformer-based models.
\end{enumerate}

This paper is organized as follows: Section 2 details the methodology and architecture of the proposed framework. Section 3 describes the experimental setup, including datasets, preprocessing, and implementation. Section 4 presents the results, with performance comparisons and an ablation study. Section 5 discusses the implications of the findings, and Section 6 concludes with a summary of contributions and future directions.

\section{Methodology}
\label{methodology}

\subsection{Task formulation}
\label{task_formu}
Our primary objective is to develop an efficient model for EEG-based abnormality detection, where we aim to accurately classify EEG signals as either normal or abnormal. 
Given a raw EEG signal 
$x \in \mathbb{R}^{C \times T}$, where $\mathbb{R}$ represents original dataset, $C$ is the number of EEG channels and $T$ denotes the temporal length of the signal. Notably, in typical EEG recordings,  $T \gg C$, emphasizing the challenge of processing long temporal sequences.

To address this, the task is divided into two stages: \textbf{latent representation learning} and \textbf{abnormality detection}. In the first stage, the raw EEG signal $\mathbf{x}$ is compressed into a compact latent representation $\mathbf{z}$ using an encoder function:
\begin{equation}
\mathbf{z} = \text{Encoder}(\mathbf{x}), \quad \mathbf{z} \in \mathbb{R}^{C \times D}
\end{equation}
where $D \ll T$, ensuring a significant reduction in computational complexity while preserving essential information. In the second stage, the compact representation $\mathbf{z}$ is input into a classifier function for final pathology prediction:
\begin{equation}
\mathbf{y} = \text{Classifier}(\mathbf{z})
\end{equation}
where $y$ represents the predicted class label (e.g., normal or abnormal EEG). Together, these stages form the backbone of the proposed CwA-T model.
\subsection{Architecture of CwA-T}
\label{structure}
\begin{figure*}[ht]
\centering
\includegraphics[width=\textwidth]{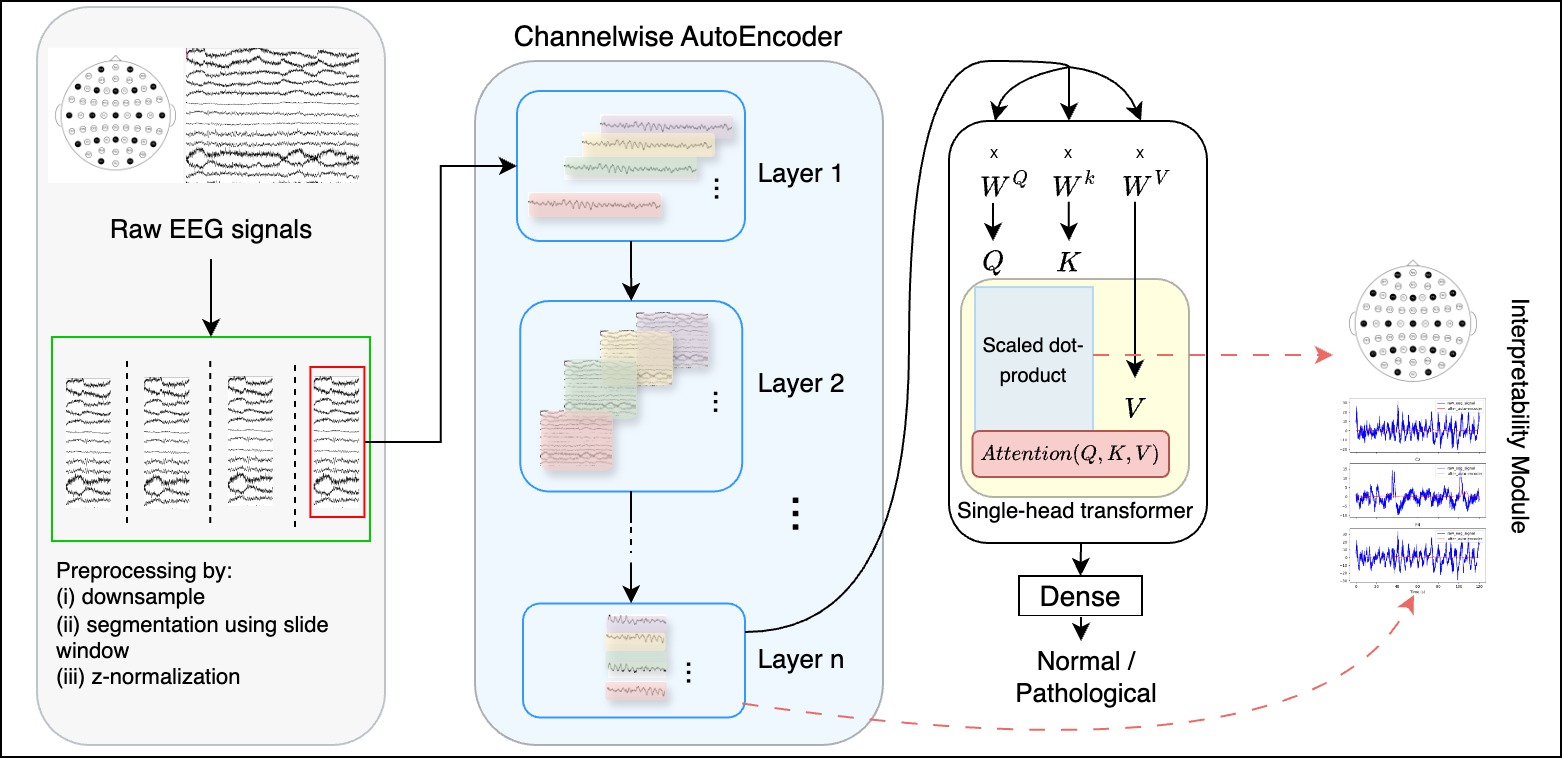}
\caption{Overview of the proposed CwA-T framework for EEG-based abnormality detection. The framework includes three primary stages: (i) Preprocessing, where raw EEG signals are downsampled, segmented using a sliding window, and z-normalized to reduce noise and standardize inputs; (ii) Channelwise Autoencoder, which compresses the preprocessed signals into latent representations through independent channelwise feature extraction, preserving the natural structure of EEG data across multiple layers; and (iii) Single-Head Transformer Classifier, which utilizes scaled dot-product attention to identify patterns in the compressed representations and outputs a classification of the signals as normal or pathological. The output also has potential to support interpretability via connectivity analysis and visualization.}\label{fig1}
\end{figure*}

The CwA-T model consists of two main components: a \textbf{channelwise autoencoder} for feature extraction and compression, and a \textbf{single-head transformer-based classifier} for abnormality detection. The overall structure is depicted in Figure  \ref{fig1}.

\subsubsection{Channelwise Autoencoder}
\label{auto-encoder}
\begin{figure}[ht]
\centering
\includegraphics[width=0.3\textwidth]{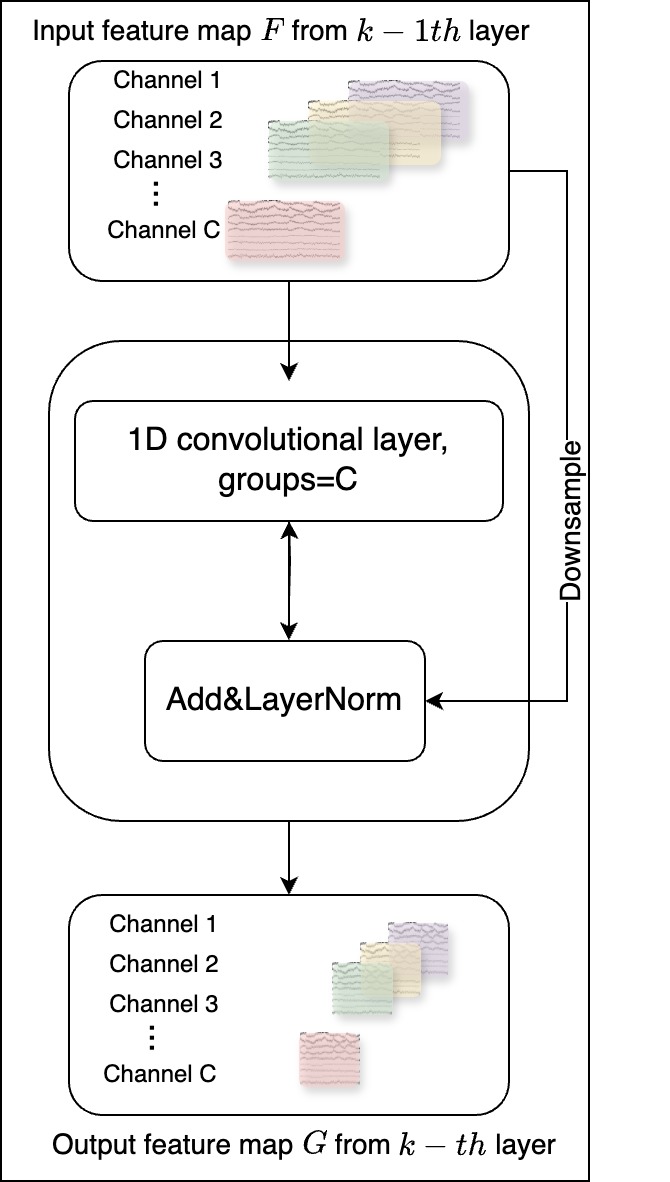}
\caption{Overview of the channelwise autoencoder architecture. The autoencoder processes input feature maps independently for each EEG channel to preserve channelwise independence. Each input feature map $F$  from the  $k-1th$ layer is passed through a 1D convolutional layer with grouped convolutions ( $\text{groups} = C$ , where  $C$  is the number of channels), followed by an addition operation and Layer Normalization ($Add\&LayerNorm$). This ensures computational efficiency and prevents inter-channel dependencies. The downsampling operation further reduces the spatial dimension of the feature maps, producing the output feature map $G$ for the $k -th$ layer. This design aligns with the natural structure of EEG data for neuroscience research and clinical applications.}\label{fig-auto-encoder}
\end{figure}

To efficiently reduce the dimensionality of raw EEG data, CwA-T employs a channelwise CNN as the core of its autoencoder. The design of our autoencoder is inspired by depthwise CNNs \cite{howard2017mobilenets, koonce2021mobilenetv3, chen2022mobile} and related architectures applied on EEG \cite{lawhern2018eegnet, schirrmeister2017deep}. While depthwise CNNs apply convolutions channel-wise to feature maps, our design treats each EEG channel independently, maintaining channelwise features without introducing inter-channel dependencies, as shown in Figure \ref{fig-auto-encoder}. This independence is crucial for preserving the natural structure of EEG data, which is essential for neuroscience research and clinical applications.

Given a input feature map $F$ with size of $D_M \times D_T$, where $D_M$ represents width of input feature map, $M$ is the number of input channels, $D_T$ is the spatial length of feature map. The output feature map $G$ for standard convolution with stride one and padding is computed as:
\begin{equation}
G_{k,j,n}=\sum_{j,m,n}=K_{j,m,n} \cdot F_{t+j-1,m}
\end{equation}
Standard convolutions have the computational cost of:
\begin{equation}
Cost_{standard}=D_K \cdot M \cdot N \cdot D_T
\end{equation}
Given the same shape of output feature map for channelwise convolution $\hat{G}$: 
\begin{equation}
\hat{G}_{k,j,n}=\sum_{j,m,n}\hat{K}_{j,\frac{m}{C},n} \cdot F_{t+j-1,m}
\end{equation}
Computational cost for channelwise convolution is:
\begin{equation}
Cost_{channelwise}=\frac{D_K \cdot M \cdot N \cdot D_T}{C}
\end{equation}
The reduction in computational cost from standard to channelwise convolution can be quantified by taking the ratio of the two costs:
\begin{equation}
\frac{Cost_{channelwise}}{Cost_{standard}}=\frac{\frac{D_K \cdot M \cdot N \cdot D_T}{C}}{D_K \cdot M \cdot N \cdot D_T}=\frac{1}{C}
\end{equation}
where $C$ represents number of EEG channels in convolutional operations.
\subsubsection{Single-Head Transformer Classifier}
\label{transformer}
The transformer architecture, introduced by Vaswani et al \cite{vaswani2017attention} in 2017, has become a foundational model in deep learning, particularly for sequence-to-sequence tasks. Its core innovation is the self-attention mechanism, which enables the model to weigh the importance of different elements in a sequence, capturing long-range dependencies without relying on recurrent structures.

The transformer consists of an encoder-decoder structure, though in many applications, such as classification tasks, only the encoder is utilized. Each encoder layer comprises two primary components: a self-attention mechanism and a position-wise fully connected feed-forward network.
\begin{enumerate}
    \item \textbf{Single-Head Self-Attention Mechanism} To improve efficiency and keep the model lightweight, we replace the multi-head setup with a single-head attention mechanism \cite{hua2022transformer}. Given an input sequence represented by the matrix $X \in \mathbb{R}^{n \times d}$ where $n$ is the sequence length and $d$ is the dimensionality of each input vector, the self-attention mechanism computes a weighted sum of the input elements, allowing the model to focus on different parts of the sequence, computed as follows:

        \textbf{Linear Projections} The input  $X$ is linearly projected to obtain single query 
        $Q$ , key $K$, and value $V$:
        \begin{equation}
        Q = XW^Q
        \end{equation}
        \begin{equation}
        K = XW^K
        \end{equation}
        \begin{equation}
        V = XW^V
        \end{equation}
        where  $W^Q, W^K, W^V \in \mathbb{R}^{d \times d_k}$ 
        are learned projection matrices, and  $d_k$  is the dimensionality of the query and key.
        
        \textbf{Scaled Dot-Product Attention} The attention score is computed by taking the dot product of the query and key, scaling by  $\sqrt{d_k}$, and applying the softmax function:
        \begin{equation}
        Attention(Q, K, V) = softmax\left( \frac{Q K^T}{\sqrt{d_k}} \right) V
        \end{equation}
        This yields a matrix of attention weights that determine the influence of each input element on the others.
        
    \item \textbf{Position-Wise Feed-Forward Network} Following the single-head attention, a position-wise feed-forward network is applied independently to each position:
    \begin{equation}
    \text{FFN}(x) = \text{ReLU}(xW_1 + b_1) W_2 + b_2
    \end{equation}
    where $W_1 \in \mathbb{R}^{d \times d_{ff}}$ and $W_2 \in \mathbb{R}^{d_{ff} \times d}$ are learned weight matrices, $b_1$ and $b_2$ are biases, and $d_{ff}$ is the dimensionality of the feed-forward layer.
    \item \textbf{Residual Connection and Layer Normalization} To facilitate training and improve convergence, residual connections and layer normalization are employed around each sub-layer:
    \begin{equation}
    \text{LayerNorm}(x + \text{Sublayer}(x))
    \end{equation}
\end{enumerate}

By employing a single head, we reduce the number of parameters significantly, leading to a more compact and computationally efficient model. The parameter count for the single-head attention mechanism is:
\begin{equation}
\text{Params}_{\text{attention}} = d \cdot d_k + d \cdot d_k + d_k \cdot d = 3 \cdot d \cdot d_k
\end{equation}
In a multi-head setup without compression, the parameter count scales with the number of heads $h$ and the full input dimensionality:
\begin{equation}
\text{Params}_{\text{multi-head attention}} = 3 \cdot d \cdot D_{orig} \cdot d_k
\end{equation}
where $D_{orig}$ represents the original signal dimensionality.

This streamlined approach, inspired by efficient transformer implementations \cite{chen2022mobile}, ensures a balance between computational efficiency and effective modeling of temporal dependencies.

\subsection{Integration of CwA-T for EEG Analysis}
The CwA-T framework integrates the channelwise autoencoder and the single-head transformer classifier into a unified model for EEG-based abnormality detection. The autoencoder first compresses the raw EEG signal into a compact latent representation, minimizing redundant information. This representation is then input into the transformer classifier, which leverages self-attention to identify patterns critical for distinguishing between normal and abnormal signals.

By combining these components, CwA-T achieves competitive performance on benchmark datasets while maintaining a lightweight design. The model’s channelwise feature extraction and efficient transformer architecture make it particularly well-suited for applications requiring both computational efficiency and interpretability.

\section{Experimental settings}
\label{experiment}
This section outlines the experimental design used to evaluate the proposed CwA-T model for EEG abnormality detection, including dataset details, preprocessing steps, implementation specifics, and the baseline models used for comparison.

\subsection{Dataset}
\label{dataset}
The TUH Abnormal EEG Corpus (v3.0.1) was utilized for experiments. This subset of the Temple University Hospital (TUH) EEG Corpus provides a balanced representation of clinically normal and abnormal EEG recordings. A total of 2,993 EEG files in European Data Format (EDF) were split into training and evaluation sets.
\begin{itemize}
    \item \textbf{Training set}: 2,717 recordings (1,371 normal, 1,346 abnormal) collected from 2,130 subjects.
    \item \textbf{Evaluation set}: 276 recordings (150 normal, 126 abnormal) involving 253 unique subjects.
\end{itemize}

Segments exceeding 15 minutes were selected for analysis to ensure high-quality data. The dataset also included diverse demographic representations, encompassing variations in age and gender, facilitating robust generalization tests for the model.

\subsection{Data preprocessing}
\label{preprocessing}
\begin{figure}[ht]
\centering
\includegraphics[width=0.45\textwidth]{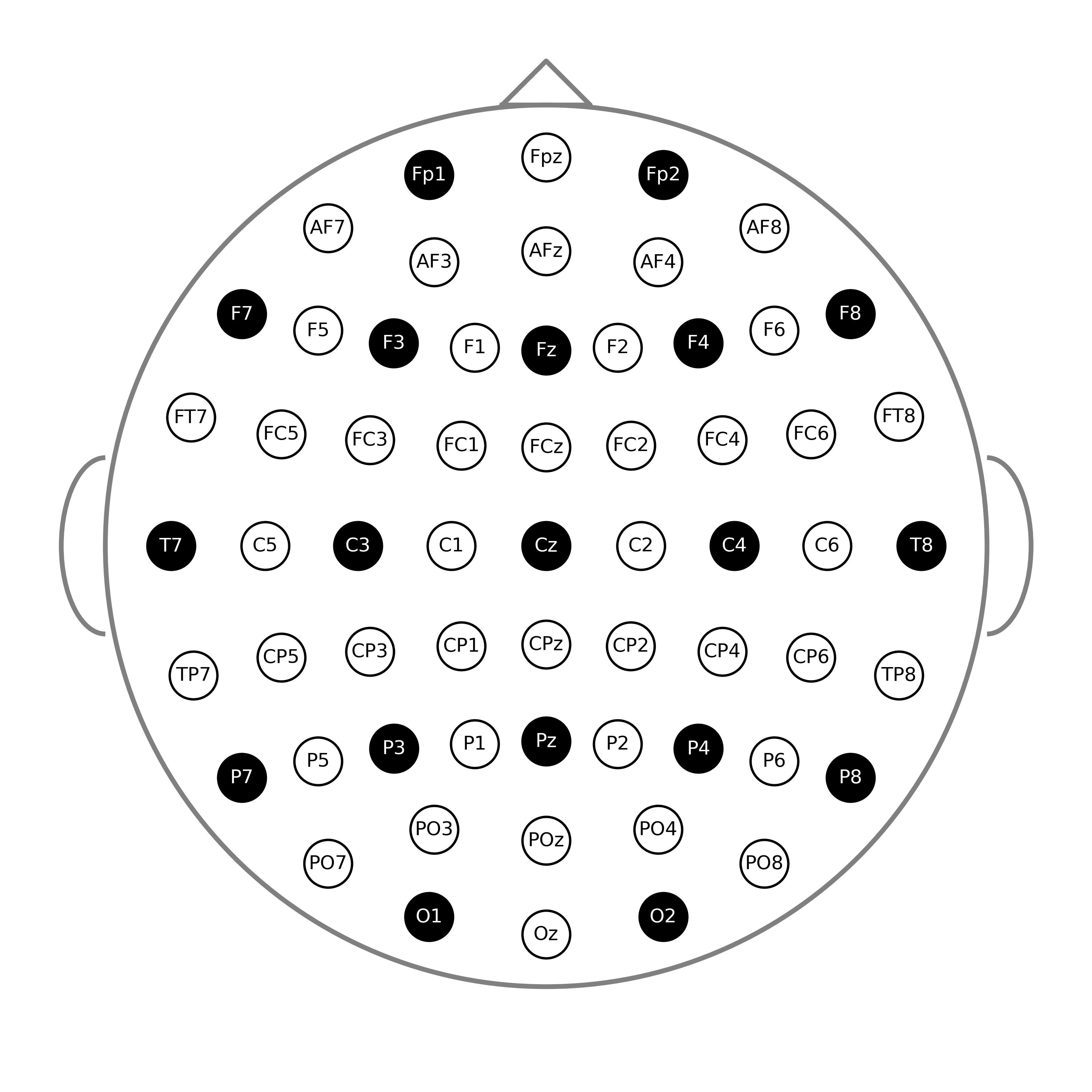}
\caption{The electrode placement in the standard 10-20 system. In our study, we use the signals from Fp1, Fp2, F3, F4, F7, F8, C3, C4, T7, T8, P3, P4, P7, P8, O1, O2, Fz, Cz, and Pz.}\label{montage}
\end{figure}

To address EEG-specific challenges, such as noise and high dimensionality, the following preprocessing steps were applied:
\begin{enumerate}
    \item \textbf{Downsampling}: EEG data was resampled to 100 Hz, maintaining sufficient resolution for frequencies up to 50 Hz (aligned with the Nyquist theorem \cite{cui2017carrier}).
    \item \textbf{Segmentation}: EEG recordings were divided into non-overlapping, 2-minute segments to standardize input lengths and optimize batch processing.
    \item \textbf{z-Normalization}: Signal amplitudes were normalized to reduce variability caused by electrode placement or individual physiology.
\end{enumerate}
The preprocessed data retained clinically relevant information across all major EEG bands ($\delta$: 0.5-4 Hz, $\theta$: 4-8 Hz, $\alpha$: 8-13 Hz, $\beta$: 13-30 Hz, $\gamma$: 30-50 Hz), supporting accurate analysis of neural activity. The EEG recordings used in this study follow the standard 10-20 system montage, as shown in Figure \ref{montage}.

\subsection{Implementation details}
\label{implementation}
For model training, we utilized the Adam optimizer with an initial learning rate of 0.001 and a weight decay of $1 \times 10^{-6}$. To promote generalization, we incorporated $L_2$-norm regularization and dropout. The batch size was set to 64 across all experiments, and training was conducted for 15 epochs.

The input signal length varied based on the segment duration, with 2-minute segments containing 12,000 data points. A warm-up phase of 200 steps was applied, during which the learning rate was gradually increased from the initial value of 0.001 to allow for stable convergence.

We divided the training and validation datasets in a 9:1 ratio, ensuring no subject overlap between the two sets. This approach avoids data leakage and allows for an unbiased evaluation of the model’s generalization capability.

Our experimental setup included Ubuntu 20.04 LTS, with PyTorch 1.12.1 and CUDA 11.3 for GPU acceleration. We ran the experiments on a single NVIDIA RTX 4090 GPU, supported by an AMD Ryzen 9 7900X CPU and 64GB of RAM, ensuring ample computational resources for efficient training and evaluation.

\subsection{Performance metrics}
\label{metrics}
To comprehensively evaluate the performance of our proposed model, we employ a set of metrics that measure both its classification performance and computational efficiency.

We use three primary metrics, sensitivity, specificity, and accuracy to evaluate the model’s ability to classify EEG signals at both the signal and case levels. These metrics are defined based on true positives (TP), true negatives (TN), false positives (FP), and false negatives (FN).

\textbf{Per-Signal Evaluation}: Each EEG signal is treated as an independent entity and the sensitivity, specificity, and accuracy are computed directly based on the model’s predictions for individual signals. This approach ensures fine-grained performance assessment and highlights the model’s ability to handle variability within signals.

\textbf{Per-Case Evaluation}: To account for clinical relevance, we aggregate predictions across all signals belonging to a single case. The case-level prediction is determined using a majority voting mechanism, where the case is classified as positive if the majority of its signals are predicted to be positive. Given the number of positive predictions ( $P$ ) among the  $N$  signals. The case is classified as positive if  $P > \frac{N}{2}$; otherwise, it is classified as negative. Sensitivity, specificity, and accuracy at the case level are then computed using the same formulas.

\begin{equation}
Sensitivity=\frac{TP}{TP+FN}
\end{equation}
\begin{equation}
Specificity=\frac{TN}{TN+FP}
\end{equation}
\begin{equation}
Accuracy=\frac{TP+TN}{TP+TN+FP+FN}
\end{equation}
To evaluate the computational efficiency of the proposed model, we further report the floating-point operations (FLOPs) and the total number of parameters. These metrics are essential for quantifying the computational complexity and scalability of the model.

\section{Results}
\label{results}
In this section, we present the performance evaluation of our proposed model for EEG-based abnormality detection. To provide a comprehensive analysis, we compare the performance of our model with several EEG-based architectures, including: EEGNet \cite{lawhern2018eegnet}, EEG-ARNN \cite{eeg-arnn}, DeepCNN \cite{schirrmeister2017deep} and FusionCNN \cite{muhammad2020eeg}.

\subsection{Subject-independent results}
\label{independent}
The results presented in Table \ref{tab:performance_comparison} compare our proposed model with several baselines, including EEGNet, EEG-ARNN, Deep4Conv, and FusionCNN, for both per-signal and per-case evaluations. These results provide a comprehensive evaluation of the strengths and weaknesses of each model.
\begin{table*}[!ht]
\centering
\caption{Performance comparison of different models on subject-independent results. Metrics include sensitivity, specificity, and accuracy for both per-signal and per-case evaluations.}
\begin{tabular}{l|ccc|ccc}
\hline
\multirow{2}{*}{Model} & \multicolumn{3}{c|}{Per-signal}    & \multicolumn{3}{c}{Per-case}     \\ \cline{2-7} 
& Sensitivity & Specificity & Accuracy & Sensitivity & Specificity & Accuracy \\ \hline
EEGNet & 98.5\% & 3.0\% & 46.9\% & 100\% & 0\% & 37.6\%   \\
EEG-ARNN & 53.5\% & 75.0\% & 65.1\% & 29.8\% & 86.8\% & 65.6\% \\
Deep4Conv & 76.3\% & 87.1\% & 82.1\% & 57.5\% & 89.9\% & 77.9\%  \\
FusionCNN & 57.0\% & 96.1\% & 78.1\% & 48.4\% & 92.6\% & 76.7\% \\ \hline
Proposed  & 72.8\% & 84.5\% & 79.1\% & 76.2\% & 91.2\% & 85.0\% \\ \hline
\end{tabular}
\label{tab:performance_comparison}
\end{table*}
EEGNet demonstrates a significant bias in its performance, with a high sensitivity of 98.5\% but an extremely low specificity of 3.0\% in the per-signal evaluation. This imbalance suggests that EEGNet struggles to generalize effectively in long-term signals, potentially due to its shallow architecture, which may not be capable of extracting robust and discriminative features. Although EEGNet is lightweight, it appears inadequate for processing long-term EEG signals, leading to biased predictions.

In the per-signal evaluation, EEG-ARNN, Deep4Conv, FusionCNN, and our proposed model all demonstrate the ability to extract discriminative features from long-term signals, achieving accuracies of 65.1\%, 82.1\%, 78.1\%, and 79.1\%, respectively. However, both EEG-ARNN and FusionCNN exhibit significant biases, with relatively high specificity but notably low sensitivity, indicating potential overfitting to certain signal characteristics. Deep4Conv achieves the highest overall performance in per-signal evaluations, outperforming our proposed model by 3.5\% in sensitivity, 2.6\% in specificity, and 3.0\% in accuracy.

In the per-case evaluation, our proposed model achieves the highest accuracy of 85.0\%, outperforming all other baselines. Furthermore, it demonstrates the best sensitivity (76.2\%) among the evaluated models. While FusionCNN shows a slightly higher specificity (92.6\%) compared to our model (91.2\%), its significantly lower sensitivity (48.4\%) highlights a critical imbalance, rendering it less reliable for real-world applications. This balance in sensitivity and specificity underscores the effectiveness of our proposed model in providing accurate and consistent predictions across diverse cases.

In summary, the proposed model demonstrates competitive performance, particularly in per-case evaluations, where it outperforms all baselines in both accuracy and sensitivity. These results underscore its ability to effectively address the challenges of long-term EEG signal processing while maintaining a balanced and reliable performance on various metrics.

\subsection{FLOPs count comparison}
\label{flops}
The results presented in Table \ref{tab_flops} provide a detailed comparison of FLOPs and the number of parameters for different models. These findings emphasize the computational advantages of our proposed model.

First, while EEGNet exhibits the lowest FLOPs (103.6M) and parameter count (19.9K) among all models, its shallow architecture limits its ability to extract meaningful features for this task, as demonstrated by its subpar classification performance. This highlights the trade-off between computational efficiency and feature extraction capability in lightweight networks.

Second, among the CNN- and GNN-based networks, including Deep4Conv, FusionCNN, and EEG-ARNN, our proposed model demonstrates competitive computational efficiency. Notably, only Deep4Conv achieves lower FLOPs (185.3M) than our model (202.0M), whereas FusionCNN, despite requiring significantly higher FLOPs (4.5G) and parameters (3.4M), does not outperform our model in classification performance. Although our model incorporates a transformer encoder, which inherently demands additional computational resources, it remains comparable to these CNN- and GNN-based baselines in terms of computational cost, showcasing its efficient design.

Lastly, we conducted a direct comparison of computational efficiency with a standalone single-head transformer. Using the same input signal configuration (19 channels with a length of 2 minutes), our model significantly reduced computational requirements compared to the single-head transformer, which requires 11.9G FLOPs and 1.3G parameters. This demonstrates that our integration of a depth-wise CNN-based auto-encoder effectively compresses the data before the transformer, leading to substantial savings in computational resources.

Overall, the proposed model achieves a commendable balance between computational efficiency and performance. By incorporating a transformer encoder, it maintains competitive FLOPs and parameter counts when compared to CNN- and GNN-based networks, while substantially reducing resource consumption relative to a standard transformer. This level of efficiency highlights its practicality and potential for real-world applications in long-term EEG signal processing.

\begin{table}[!htbp]
\centering
\caption{Comparison of FLOPs and parameters across different models}
\begin{tabular}{l c c}
  Network & FLOPs & Params \\  \hline
  EEGNet & 103.6M & 19.9K \\
  EEG-ARNN & 260.0M & 69.8K \\
  Deep4Conv & 185.3M & 0.2M \\
  FusionCNN & 4.5G & 3.4G \\ \hline
  Single-head transformer$^*$ & 11.9G & 1.3G \\ \hline
  Proposed & 202.0M & 2.9M \\ \hline
\end{tabular}
\vspace{2mm} 
\noindent
\caption*{$*$ The single-head transformer model was not trained or evaluated for performance, only the FLOPs and parameter count were calculated for reference purposes.}
\label{tab_flops}
\end{table}


\subsection{Ablation study}
\label{ablation}
\begin{figure}[!htbp]
    \centering
    \begin{subfigure}{0.49\textwidth}  
        \centering
        \includegraphics[width=\textwidth]{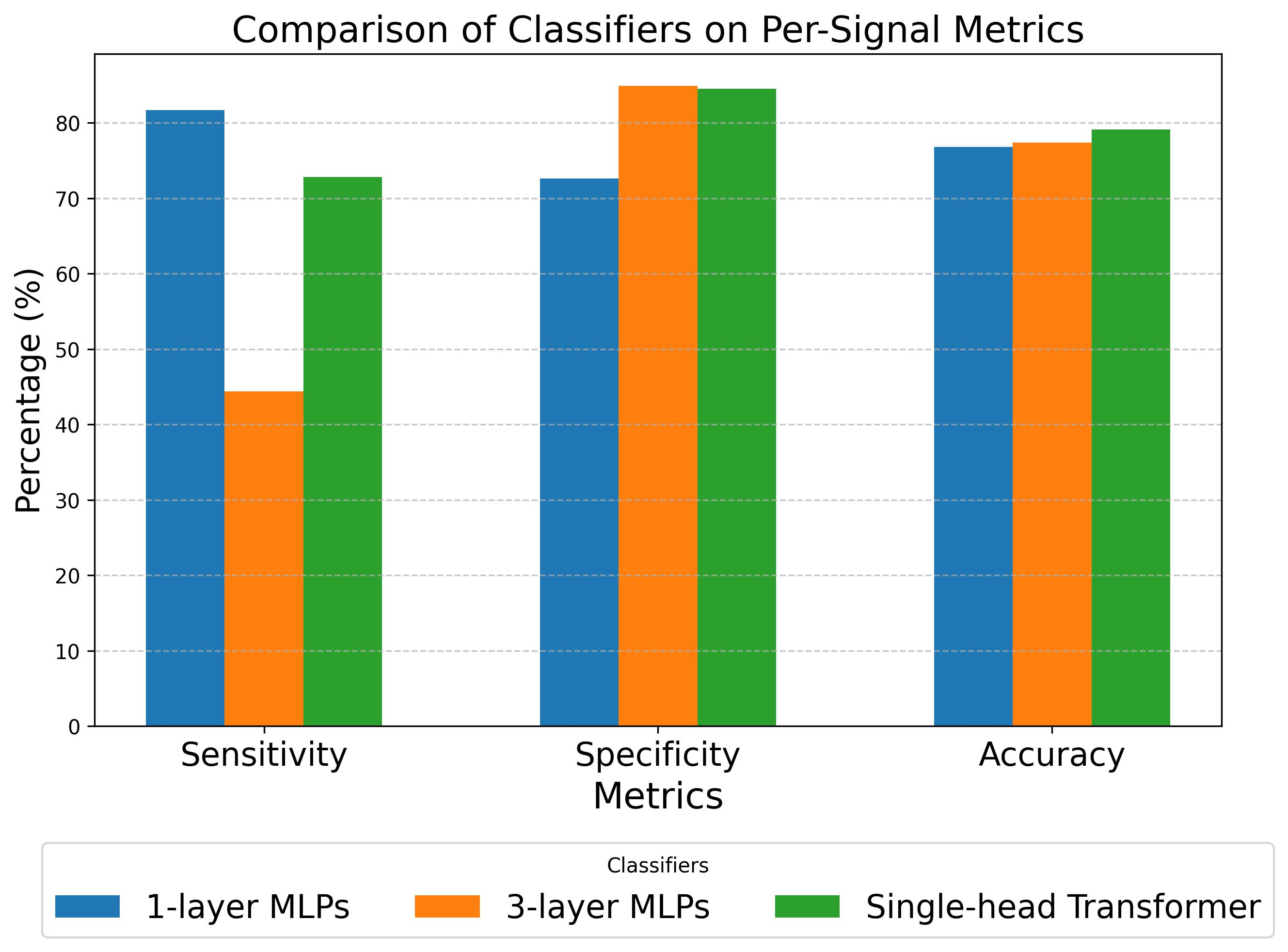}  
        \caption{Per-signal comparison}
        \label{mlp_a}
    \end{subfigure}
    \hfill
    \begin{subfigure}{0.49\textwidth}  
        \centering
        \includegraphics[width=\textwidth]{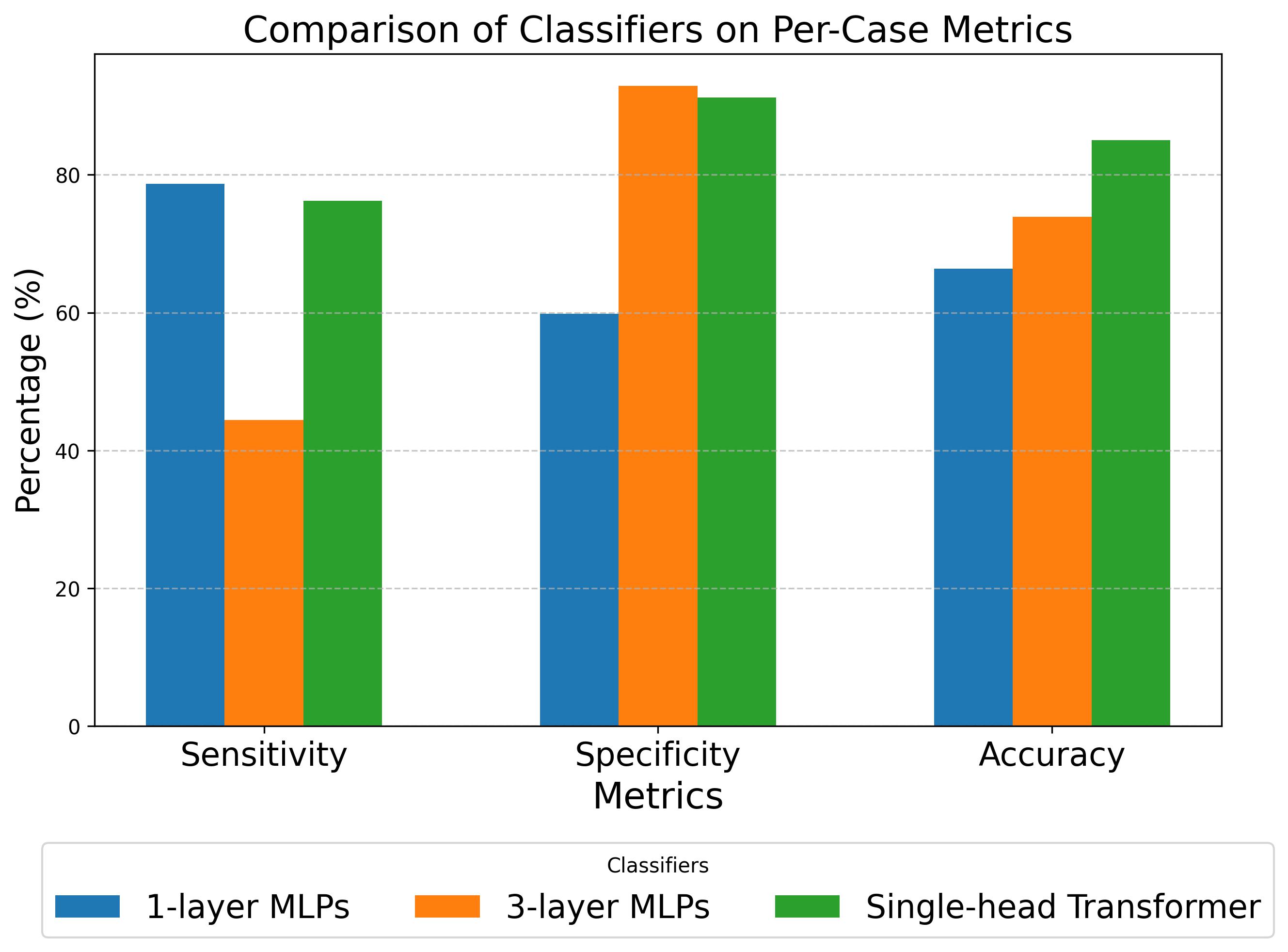}  
        \caption{Per-case comparison}
        \label{mlp_b}
    \end{subfigure}
    
    \caption{Performance comparison of classifiers between MLPs and Single-head transformer}
    \label{mlp}
\end{figure}

\textbf{Single-head transformer-encoder vs. MLPs}
Multi-head attention-based transformer architectures have been extensively studied and widely adopted for application. The single-head transformer-encoder is designed to process EEG signals effectively while maintaining a lightweight structure. To evaluate its performance, we compare it with per-signal and per-case metrics (see Figure \ref{mlp}). Furthermore, we believe that transformer-encoder has great potential for advancing neuroscience research, particularly in applications requiring high performance and interpretable models.

\textbf{LayerNorm in autoencoder} The design of our autoencoder intentionally avoids creating new features among EEG channels to preserve their independence. This channelwise independence is critical for maintaining the natural structure of EEG signals, which is essential for neuroscience research and clinical applications. Unlike BatchNorm, which normalizes features across the batch dimension and inherently introduces inter-channel dependencies, it is necessary to use a LayerNorm method that operates independently across channels. Also, LayerNorm provides a more reasonable and interpretable approach for EEG-related studies. For example, in EEG signals, certain channels such as Fp1 and Fp2 (located near the forehead) are more prone to artifacts from eye blinks and eye movements, whereas posterior channels such as O1 and O2 are less affected by such artifacts. LayerNorm normalizes each channel independently, preventing the transfer of artifact-related statistics (mean and variance) between channels.

\section{Discussion}
\label{discussion}
In this section, we discuss the contributions and implications of our proposed model for EEG-based abnormality detection. Key aspects include computational efficiency achieved through the CNN-based auto-encoder, the interpretability of the architecture, and its ability to handle long-term EEG signals effectively.
\subsection{Trade-offs between performance and efficiency}
\label{tradeoffs}
The proposed CwA-T framework achieves an optimal trade-off between computational efficiency and classification performance, addressing the inherent challenges of long-term EEG signal processing. Compared to lightweight models such as EEGNet, which are computationally inexpensive but exhibit biased predictions, our model introduces additional computational cost justified by substantial performance improvements. Specifically, CwA-T achieves an accuracy of 85.0\% at the per-case level while maintaining competitive sensitivity and specificity.

The channelwise autoencoder plays a critical role in ensuring this efficiency. By compressing raw EEG signals into compact representations, it significantly reduces the input size for the transformer-based classifier. This compression alleviates the computational burden associated with transformer models while retaining essential spatial and temporal information. For instance, our framework requires only 202M FLOPs, a drastic reduction compared to the standalone single-head transformer model, which requires 11.9G FLOPs.

The single-head transformer classifier further enhances the framework’s efficiency. While conventional multi-head attention mechanisms often lead to computationally heavy models, our single-head design retains the ability to model long-range dependencies effectively without introducing excessive parameters. Together, the autoencoder and lightweight transformer ensure that CwA-T remains practical for real-world EEG applications, including resource-constrained environments.
\subsection{Interpretability of the Proposed Model}
\label{discussion_interpretability}
A key strength of the CwA-T framework lies in its interpretability, which is particularly valuable for clinical and neuroscience applications. The channelwise autoencoder not only reduces data dimensionality but also preserves critical spatial and temporal features, enabling insights into neural dynamics and abnormal brain activity.

\begin{figure}[!htbp]
    \centering
    \begin{subfigure}{0.45\textwidth}  
        \centering
        \includegraphics[width=\textwidth]{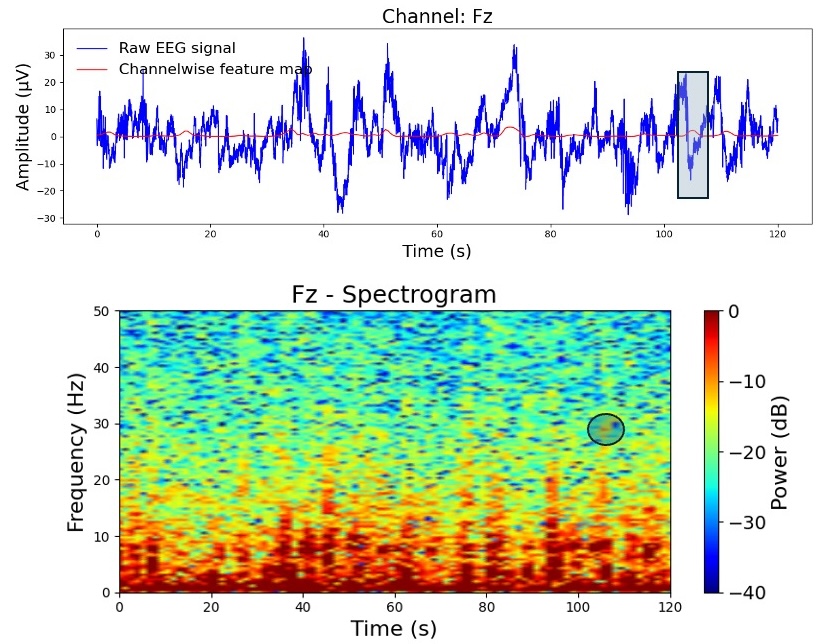}  
        \caption{Channel Fz: An anomaly is observed near the 105-second mark, where the autoencoder output reveals elevated beta activity. The corresponding spectrogram highlights an increased power around 30 Hz.}
        \label{feature_map_fz}
    \end{subfigure}
    \hfill
    \begin{subfigure}{0.45\textwidth}  
        \includegraphics[width=\textwidth]{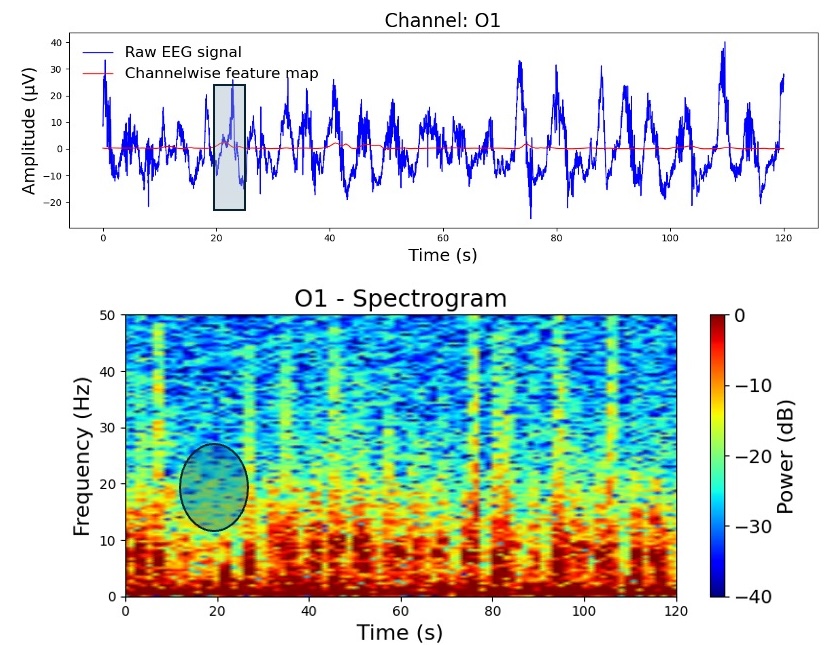}  
        \caption{Channel O1: A deficiency in alpha activity (8–13 Hz) is detected in the occipital region. The autoencoder output reflects a reduction in signal amplitude, confirmed by the absence of spectral power in the alpha band in the spectrogram.}
        \label{feature_map_o1}
    \end{subfigure}
    \vspace{0.5cm}
    \begin{subfigure}{0.45\textwidth}  
        \centering
        \includegraphics[width=\textwidth]{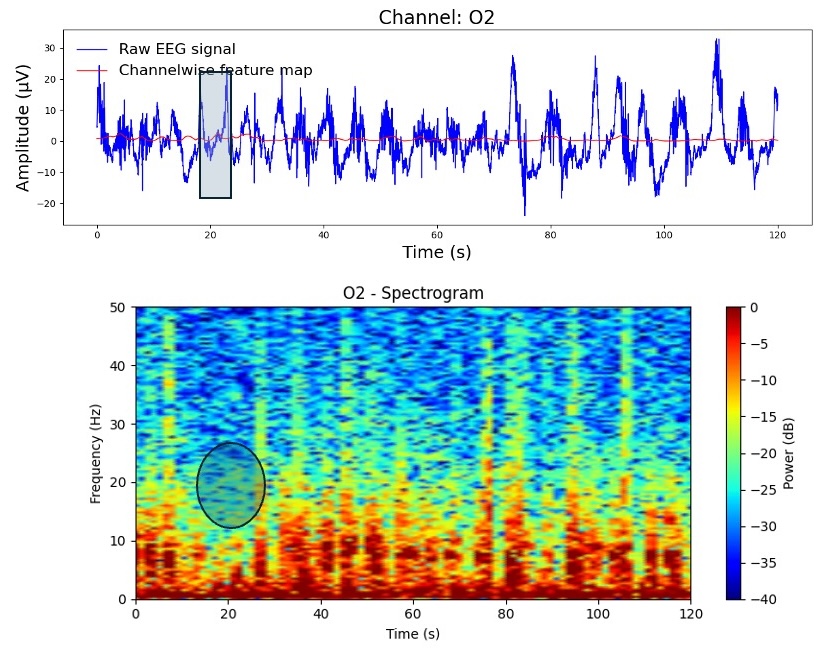}  
        \caption{Channel O2: Similar to Channel O1, a deficiency in alpha activity is observed, where both the autoencoder output and spectrogram highlight the lack of prominent alpha rhythms.}
        \label{feature_map_o2}
    \end{subfigure}
    
    \caption{\textbf{Examples of EEG signals with annotated abnormalities and corresponding autoencoder outputs.} This figure illustrates samples of EEG signals from abnormal patients alongside the outputs of the channelwise autoencoder. The raw EEG signal (blue line), autoencoder output (red line), and corresponding spectrograms are shown for 2-minute segments. Anomalous points, identified by rectangles and circles, suggest potential abnormalities detected by the model.}
    \label{feature_map}
\end{figure}

To demonstrate this, we analyzed the outputs of the autoencoder alongside raw EEG signals and corresponding spectrograms. Two representative examples from pathological EEG signals highlight the model’s ability to retain biologically interpretable features:
\begin{enumerate}
    \item \textbf{Abnormal Beta Activity}: Beta rhythms (13–30 Hz) are typically sparse in healthy scalp EEG recordings and rarely exceed 25 Hz. In one pathological signal, the autoencoder identified an abnormal beta activity pattern around the 105-second mark. This anomaly was confirmed by the spectrogram of channel Fz (see Figure \ref{feature_map_fz}), which displayed elevated spectral power around 30 Hz. Such results emphasize the model’s ability to detect subtle frequency-domain abnormalities that align with known pathological EEG characteristics.
    \item \textbf{Deficiency in Alpha Activity}: Alpha waves (8–13 Hz) primarily originate in the occipital lobe and exhibit amplitudes between 15–45$\mu$V. In a second example, the autoencoder output revealed a significant absence of alpha activity in channels O1 and O2, corroborating the lack of alpha rhythms observed in the raw EEG signals and corresponding spectrograms, as shown in Figure \ref{feature_map_o1} and Figure \ref{feature_map_o2}. Since alpha wave disruptions are often indicative of neurological disorders, such findings validate the model’s potential for identifying spatially relevant abnormalities.
\end{enumerate}

These examples underscore the interpretability of the channelwise autoencoder, which bridges the gap between black-box deep learning models and traditional neuroscience approaches. By maintaining channel independence, the autoencoder facilitates biologically meaningful analysis of EEG signals while enabling region-specific anomaly detection.

Future investigations into the interpretability of the transformer-based classifier hold significant promise. In future work, we aim to analyze the similarity matrices generated by the self-attention mechanism, as these matrices provide insights into the relationships between EEG channels. Such analyses may uncover functional brain connectivity patterns, further enhancing the clinical utility of our framework.

\subsection{Capability for Long-term EEG signal processing}
\label{dis_long-term}
Long-term EEG signals, often exceeding 2 minutes, pose unique challenges due to their size, complexity, and susceptibility to noise. Traditional deep learning approaches often struggle with such sequences, either due to computational limitations or an inability to retain relevant temporal information. Our proposed CwA-T framework effectively addresses these issues by leveraging the channelwise autoencoder to compress long sequences into compact latent representations while maintaining essential features.

The experimental results demonstrate the model’s robustness in processing extended EEG recordings. CwA-T achieves high accuracy and balanced sensitivity-specificity performance across both per-signal and per-case evaluations. These findings suggest that the framework can generalize effectively to long-term EEG signals, a critical requirement for clinical applications such as continuous brain monitoring and diagnosis of neurological disorders.

Moreover, the ability of model to handle long sequences without loss of interpretability expands its applicability to real-time EEG analysis in clinical settings. For instance, continuous EEG monitoring for seizure detection or neurological assessment often generates large volumes of data that require efficient processing. CwA-T’s lightweight design and interpretability make it a strong candidate for deployment in such scenarios, offering both computational efficiency and clinically relevant outputs.
\subsection{Future Directions}
\label{dis_future_work}
The results of this study highlight the potential of CwA-T to transform EEG-based abnormality detection and neural connectivity analysis. By combining computational efficiency, interpretability, and competitive performance, the framework provides a robust solution to challenges in long-term EEG signal analysis.

In future work, we aim to:
\begin{enumerate}
    \item \textbf{Explore the transformer’s similarity matrices} to identify functional brain connectivity patterns and validate these findings against established neuroscientific theories.
    \item \textbf{Extend the framework} to handle multi-modal EEG datasets, integrating complementary modalities such as fMRI or MEG for more comprehensive neural analysis.
    \item \textbf{Investigate additional applications} of the channelwise autoencoder in neuroscience, such as region-specific feature extraction for studying brain disorders like epilepsy, Alzheimer’s disease, and Parkinson’s disease.
\end{enumerate}
By addressing these areas, we hope to further enhance the clinical relevance and impact of our model in neuroscience research and medical diagnostics.
\section{Conclusion}
\label{conclusion}
This study introduced CwA-T, a novel framework for EEG-based abnormality detection that integrates a channelwise CNN-based autoencoder with a transformer-based classifier. By addressing key challenges in long-term EEG signal analysis, CwA-T demonstrates a balance between computational efficiency, interpretability, and high classification performance.

The channelwise autoencoder effectively reduces the dimensionality of raw EEG signals while preserving channel independence, facilitating biologically meaningful feature extraction. This design ensures that the model outputs remain interpretable, providing insights into spatial and temporal patterns critical for clinical applications. The transformer-based classifier further leverages the compressed representations, achieving superior accuracy and sensitivity compared to baseline models, such as EEGNet, Deep4Conv, and FusionCNN, while maintaining competitive specificity.

CwA-T has demonstrated its ability to handle long-term EEG signals efficiently, achieving a substantial reduction in computational cost relative to standalone transformer models. Its lightweight architecture and interpretable outputs make it a practical and reliable tool for real-world applications, such as continuous monitoring in clinical settings or large-scale neuroscience studies.

Looking ahead, future work will focus on expanding the interpretability of CwA-T through analysis of self-attention similarity matrices, enabling deeper insights into functional brain connectivity. Additional efforts will explore the integration of CwA-T with multi-model datasets and its application in analyzing specific neurological disorders.

In summary, the proposed CwA-T framework provides a robust, interpretable, and efficient solution for EEG signal analysis. It paves the way for advancements in brain abnormality detection and opens new opportunities for neuroscience research and clinical diagnostics.

\section*{Declaration of Competing Interest}
The authors declare that they have no known competing financial interests or personal relationships that could have appeared to influence the work reported in this paper.

\section*{Data Availability}
Data will be made available on request.

\section*{Acknowledgment}
The authors thank Xiaoliang Wu, Haobo Zhou and Chenyang Lin, Masato Kudo for providing valuable suggestions that helped improve this paper.
\bibliographystyle{unsrt}  
\bibliography{ref}

\begin{thebibliography}{10}

\bibitem{feigin2019global}
Valery~L Feigin, Emma Nichols, Tahiya Alam, Marlena~S Bannick, Ettore Beghi, Natacha Blake, William~J Culpepper, E~Ray Dorsey, Alexis Elbaz, Richard~G Ellenbogen, et~al.
\newblock Global, regional, and national burden of neurological disorders, 1990--2016: a systematic analysis for the global burden of disease study 2016.
\newblock {\em The Lancet Neurology}, 18(5):459--480, 2019.

\bibitem{feigin2021burden}
Valery~L Feigin, Theo Vos, Fares Alahdab, Arianna Maever~L Amit, Till~Winfried B{\"a}rnighausen, Ettore Beghi, Mahya Beheshti, Prachi~P Chavan, Michael~H Criqui, Rupak Desai, et~al.
\newblock Burden of neurological disorders across the us from 1990-2017: a global burden of disease study.
\newblock {\em JAMA neurology}, 78(2):165--176, 2021.

\bibitem{rivera2022diagnosis}
Manuel~J Rivera, Miguel~A Teruel, Alejandro Mate, and Juan Trujillo.
\newblock Diagnosis and prognosis of mental disorders by means of eeg and deep learning: a systematic mapping study.
\newblock {\em Artificial Intelligence Review}, pages 1--43, 2022.

\bibitem{lima2022comprehensive}
Aklima~Akter Lima, M~Firoz Mridha, Sujoy~Chandra Das, Muhammad~Mohsin Kabir, Md~Rashedul Islam, and Yutaka Watanobe.
\newblock A comprehensive survey on the detection, classification, and challenges of neurological disorders.
\newblock {\em Biology}, 11(3):469, 2022.

\bibitem{dara2018feature}
Suresh Dara and Priyanka Tumma.
\newblock Feature extraction by using deep learning: A survey.
\newblock In {\em 2018 Second international conference on electronics, communication and aerospace technology (ICECA)}, pages 1795--1801. IEEE, 2018.

\bibitem{amin2015feature}
Hafeez~Ullah Amin, Aamir~Saeed Malik, Rana~Fayyaz Ahmad, Nasreen Badruddin, Nidal Kamel, Muhammad Hussain, and Weng-Tink Chooi.
\newblock Feature extraction and classification for eeg signals using wavelet transform and machine learning techniques.
\newblock {\em Australasian physical \& engineering sciences in medicine}, 38:139--149, 2015.

\bibitem{sun2023survey}
Congzhong Sun and Chaozhou Mou.
\newblock Survey on the research direction of eeg-based signal processing.
\newblock {\em Frontiers in Neuroscience}, 17:1203059, 2023.

\bibitem{hamad2016feature}
Asmaa Hamad, Essam~H Houssein, Aboul~Ella Hassanien, and Aly~A Fahmy.
\newblock Feature extraction of epilepsy eeg using discrete wavelet transform.
\newblock In {\em 2016 12th international computer engineering conference (ICENCO)}, pages 190--195. IEEE, 2016.

\bibitem{panda2010classification}
Rajanikant Panda, PS~Khobragade, PD~Jambhule, SN~Jengthe, PR~Pal, and TK~Gandhi.
\newblock Classification of eeg signal using wavelet transform and support vector machine for epileptic seizure diction.
\newblock In {\em 2010 International conference on systems in medicine and biology}, pages 405--408. IEEE, 2010.

\bibitem{velasco2023motor}
Ivan Velasco, A~Sipols, C~Simon De~Blas, Luis Pastor, and Sofia Bayona.
\newblock Motor imagery eeg signal classification with a multivariate time series approach.
\newblock {\em BioMedical Engineering OnLine}, 22(1):29, 2023.

\bibitem{wang2018epileptic}
Dong Wang, Doutian Ren, Kuo Li, Yiming Feng, Dan Ma, Xiangguo Yan, and Gang Wang.
\newblock Epileptic seizure detection in long-term eeg recordings by using wavelet-based directed transfer function.
\newblock {\em IEEE Transactions on Biomedical Engineering}, 65(11):2591--2599, 2018.

\bibitem{song2022eeg}
Yonghao Song, Qingqing Zheng, Bingchuan Liu, and Xiaorong Gao.
\newblock Eeg conformer: Convolutional transformer for eeg decoding and visualization.
\newblock {\em IEEE Transactions on Neural Systems and Rehabilitation Engineering}, 31:710--719, 2022.

\bibitem{xie2022transformer}
Jin Xie, Jie Zhang, Jiayao Sun, Zheng Ma, Liuni Qin, Guanglin Li, Huihui Zhou, and Yang Zhan.
\newblock A transformer-based approach combining deep learning network and spatial-temporal information for raw eeg classification.
\newblock {\em IEEE Transactions on Neural Systems and Rehabilitation Engineering}, 30:2126--2136, 2022.

\bibitem{wei2023tc}
Yi~Wei, Yu~Liu, Chang Li, Juan Cheng, Rencheng Song, and Xun Chen.
\newblock Tc-net: A transformer capsule network for eeg-based emotion recognition.
\newblock {\em Computers in biology and medicine}, 152:106463, 2023.

\bibitem{lih2023epilepsynet}
Oh~Shu Lih, V~Jahmunah, Elizabeth~Emma Palmer, Prabal~D Barua, Sengul Dogan, Turker Tuncer, Salvador Garc{\'\i}a, Filippo Molinari, and U~Rajendra Acharya.
\newblock Epilepsynet: Novel automated detection of epilepsy using transformer model with eeg signals from 121 patient population.
\newblock {\em Computers in Biology and Medicine}, 164:107312, 2023.

\bibitem{xu2020one}
Gaowei Xu, Tianhe Ren, Yu~Chen, and Wenliang Che.
\newblock A one-dimensional cnn-lstm model for epileptic seizure recognition using eeg signal analysis.
\newblock {\em Frontiers in neuroscience}, 14:578126, 2020.

\bibitem{supakar2022deep}
Rinku Supakar, Parthasarathi Satvaya, and Prasun Chakrabarti.
\newblock A deep learning based model using rnn-lstm for the detection of schizophrenia from eeg data.
\newblock {\em Computers in Biology and Medicine}, 151:106225, 2022.

\bibitem{li2022eeg}
Chang Li, Zhongzhen Zhang, Xiaodong Zhang, Guoning Huang, Yu~Liu, and Xun Chen.
\newblock Eeg-based emotion recognition via transformer neural architecture search.
\newblock {\em IEEE Transactions on Industrial Informatics}, 19(4):6016--6025, 2022.

\bibitem{adadi2018peeking}
Amina Adadi and Mohammed Berrada.
\newblock Peeking inside the black-box: a survey on explainable artificial intelligence (xai).
\newblock {\em IEEE access}, 6:52138--52160, 2018.

\bibitem{howard2017mobilenets}
Andrew~G Howard, Menglong Zhu, Bo~Chen, Dmitry Kalenichenko, Weijun Wang, Tobias Weyand, Marco Andreetto, and Hartwig Adam.
\newblock Mobilenets: efficient convolutional neural networks for mobile vision applications (2017).
\newblock {\em arXiv preprint arXiv:1704.04861}, 126, 2017.

\bibitem{koonce2021mobilenetv3}
Brett Koonce and Brett Koonce.
\newblock Mobilenetv3.
\newblock {\em Convolutional Neural Networks with Swift for Tensorflow: Image Recognition and Dataset Categorization}, pages 125--144, 2021.

\bibitem{chen2022mobile}
Yinpeng Chen, Xiyang Dai, Dongdong Chen, Mengchen Liu, Xiaoyi Dong, Lu~Yuan, and Zicheng Liu.
\newblock Mobile-former: Bridging mobilenet and transformer.
\newblock In {\em Proceedings of the IEEE/CVF conference on computer vision and pattern recognition}, pages 5270--5279, 2022.

\bibitem{lawhern2018eegnet}
Vernon~J Lawhern, Amelia~J Solon, Nicholas~R Waytowich, Stephen~M Gordon, Chou~P Hung, and Brent~J Lance.
\newblock Eegnet: a compact convolutional neural network for eeg-based brain--computer interfaces.
\newblock {\em Journal of neural engineering}, 15(5):056013, 2018.

\bibitem{schirrmeister2017deep}
Robin~Tibor Schirrmeister, Jost~Tobias Springenberg, Lukas Dominique~Josef Fiederer, Martin Glasstetter, Katharina Eggensperger, Michael Tangermann, Frank Hutter, Wolfram Burgard, and Tonio Ball.
\newblock Deep learning with convolutional neural networks for eeg decoding and visualization.
\newblock {\em Human brain mapping}, 38(11):5391--5420, 2017.

\bibitem{vaswani2017attention}
A~Vaswani.
\newblock Attention is all you need.
\newblock {\em Advances in Neural Information Processing Systems}, 2017.

\bibitem{hua2022transformer}
Weizhe Hua, Zihang Dai, Hanxiao Liu, and Quoc Le.
\newblock Transformer quality in linear time.
\newblock In {\em International conference on machine learning}, pages 9099--9117. PMLR, 2022.

\bibitem{cui2017carrier}
Can Cui, Wen Wu, and Wen-Qin Wang.
\newblock Carrier frequency and doa estimation of sub-nyquist sampling multi-band sensor signals.
\newblock {\em IEEE sensors journal}, 17(22):7470--7478, 2017.

\bibitem{eeg-arnn}
Biao Sun, Zhengkun Liu, Zexu Wu, Chaoxu Mu, and Ting Li.
\newblock Graph convolution neural network based end-to-end channel selection and classification for motor imagery brain–computer interfaces.
\newblock {\em IEEE Transactions on Industrial Informatics}, 19(9):9314--9324, 2023.

\bibitem{muhammad2020eeg}
Ghulam Muhammad, M~Shamim Hossain, and Neeraj Kumar.
\newblock Eeg-based pathology detection for home health monitoring.
\newblock {\em IEEE Journal on Selected Areas in Communications}, 39(2):603--610, 2020.

\end{thebibliography}

\end{document}